\documentclass[]{article}
\usepackage[utf8]{inputenc}
\usepackage{graphicx}
\usepackage[margin=1.5in]{geometry}
\usepackage{amsmath}
\usepackage[dvipsnames]{xcolor}
\graphicspath{{images}}
\title{Transformer Based Geocoding}

\newcommand*\samethanks[1][\value{footnote}]{\footnotemark[#1]}
\author{ {\hspace{1mm}Yuval Solaz}\thanks{Equal Contribution}\\
	\texttt{yuval.solaz@gmail.com} \\
	\and
	{\hspace{1mm}Vitaly Shalumov}\samethanks \\
	\texttt{vitaly.shalumov@gmail.com} \\
}
\usepackage{resizegather} 

\renewcommand{\eqref}[1]{Eq.~(\ref{#1})}

\newcommand{\figref}[1]{Fig.~\ref{#1}}

\begin{document}
\maketitle
\begin{abstract}
In this paper, we formulate the problem of predicting a geolocation from free text as a sequence-to-sequence problem. Using this formulation, we obtain a geocoding model by training a T5 encoder-decoder transformer model using free text as an input and geolocation as an output.
The geocoding model was trained on geo-tagged wikidump data with adaptive cell partitioning for the geolocation representation. All of the code including Rest-based application, dataset and model checkpoints used in this work are publicly available.

\end{abstract}

\section{Introduction}
Social media such as Twitter and Wikipedia contains considerable amount of location-related text data.
In this paper, we develop a model that learns to predict spatial probabilities from free text. 
Given a query sentence, the model outputs a discrete probability distribution over the surface earth, by assigning each geographical cell a likelihood that the input text relates to the location inside said cell. The resulting model is capable of localizing a large variety of sentences.
Viewing the task as a hierarchical classification problem allows the model to express its uncertainty in the location associated with the text. 
The resulting model can be used for resolving ambiguity of the location references in the text.
This capability is central to the success of finding exact location from free text.  
For example, \textit{Paris} can refer to more than one possible location. In a context such as: \textit{The International Olympic Committee confirmed the city chosen to host the Olympic Games in 2024. The Games will be held in Paris}, geocoding models like the one proposed in this paper can help in the resolution of the correct location. 

This work introduces the following contributions:

\begin{itemize}
    \item Synthesizing a dataset for supervised learning, including adaptive cell partitioning.
     \item Formulating the geocoding problem as a sequence-to-sequence problem.

     \item Training an end-to-end geocoding model using said formulation. 
       \item Publicly releasing the curated dataset, a Rest-based application and the T5 geocoding model.

\end{itemize}

\section{Related works}
The geolocation prediction from free text was extensively addressed in the literature.
The authors of \cite{kulkarni2020spatial} presented a multi-level geocoding model that learns to associate texts to geographic locations. The downstream task was formulated as a multi-level classification problem based on multi-level S2 \cite{s2} cells as the output space from a multi-headed model.
In \cite{weyand2016planet}, the surface of the earth was subdivided into thousands of multi-scale geographic cells. The authors trained a deep neural network using millions of geo-tagged images. We adopt the adaptive partition approach introduced in this paper.
The authors of  \cite{kinsella2011m} created language models of locations using coordinates extracted from geo-tagged Twitter data. The locations were modeled at varying levels of granularity, from zip code to the country level.
In \cite{radford2021regressing}, an end-to-end probabilistic model for geocoding text data was presented. In addition, the authors collected a novel data set for evaluating the performance of geocoding systems. The model-based solution, called ELECTRo-map was compared to the open source system available at the time of publication for geocoding texts for event data. 
An algorithm for estimating a distribution over geographic locations from a single image using a purely data-driven scene matching approach was provided in \cite{hays2008im2gps}.

The aforementioned approaches to geocoding can be viewed as sequential tasks of token classification  (which is assumed to be solved perfectly) and geocoding. 
Unfortunately, this approach suffers from several drawbacks. 
First, it cannot handle correctly inputs that contain several descriptions of locations (for example: "We live in country X city Y").
Second, it lacks the context of free text (for example: "Country south of France").

\section{Text Geolocation with Transformer}
We pose the task of text geolocation as a sequence-to-sequence problem. The model translates input text into hierarchy set of geographic cells which represent probability distribution over the surface of the earth. The model output is a sequence encoding of hierarchical cell representation of the surface of the earth.
\subsection{Adaptive Cell Partitioning}
\begin{figure} [!ht]
\centering
\includegraphics[width=12cm]{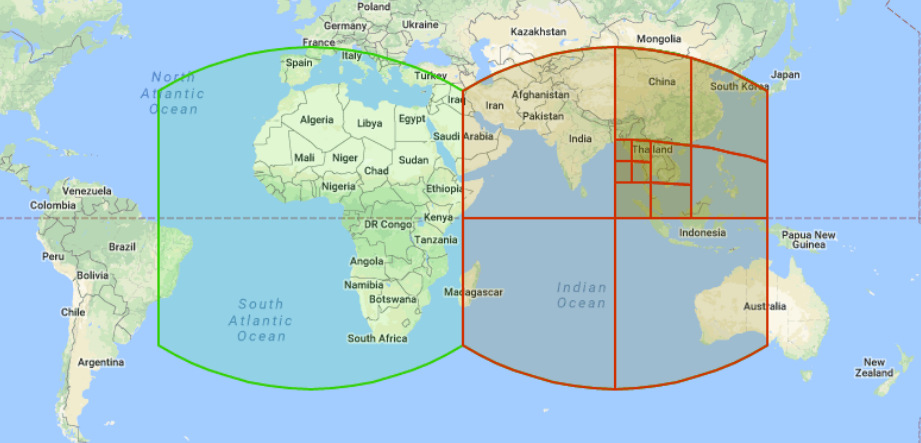}
\caption{S2 Geometry hierarchical partitioning of the earth.}
\label{fig:s2geometry}
\end{figure}

We use Google’s open source S2 Geometry library \cite{s2} to partition the surface of the earth into non-overlapping cells that define the target of our model. The S2 library defines hierarchical partitioning of the surface of a sphere by projecting the surfaces of an enclosing cube on it. The six sides of the cube are subdivided hierarchically by six quad-trees. A node in a quad-tree defines a region on the sphere called an S2 cell. \figref{fig:s2geometry} illustrates the S2 cells in several resolutions. Tab. \ref{s2geo table} shows the resolution and number of cells for each S2 Geometry level.  

\begin{table}[!ht]
\caption{S2 Geometry levels.}\label{s2geo table}
\centering
\begin{tabular}{llll}
\hline
\hline
Level& Average area& Number of cells&  \\\hline
00& 85M \(km^2\)& 6\\ 
01& 21M \(km^2\)& 24\\
02& 5M \(km^2\)& 96\\
03& 1.3M \(km^2\)& 384\\
04& 330K \(km^2\)& 1536\\
05& 83K \(km^2\)& 6K\\
06& 20K \(km^2\)& 24K\\
07& 5K \(km^2\)& 98K\\
08& 1297 \(km^2\)& 393K\\
09& 324 \(km^2\)& 1573\\
10& 81 \(km^2\)& 6M\\
..&..&..\\
29& 2.95 \(cm^2\)& \(1729*10^{15}\)\\
30& 0.74 \(cm^2\)& \(7*10^{18}\)\\
\hline
\hline
\end{tabular}
\end{table}

\begin{figure}[!htb]
\centering
\includegraphics[width=13cm]{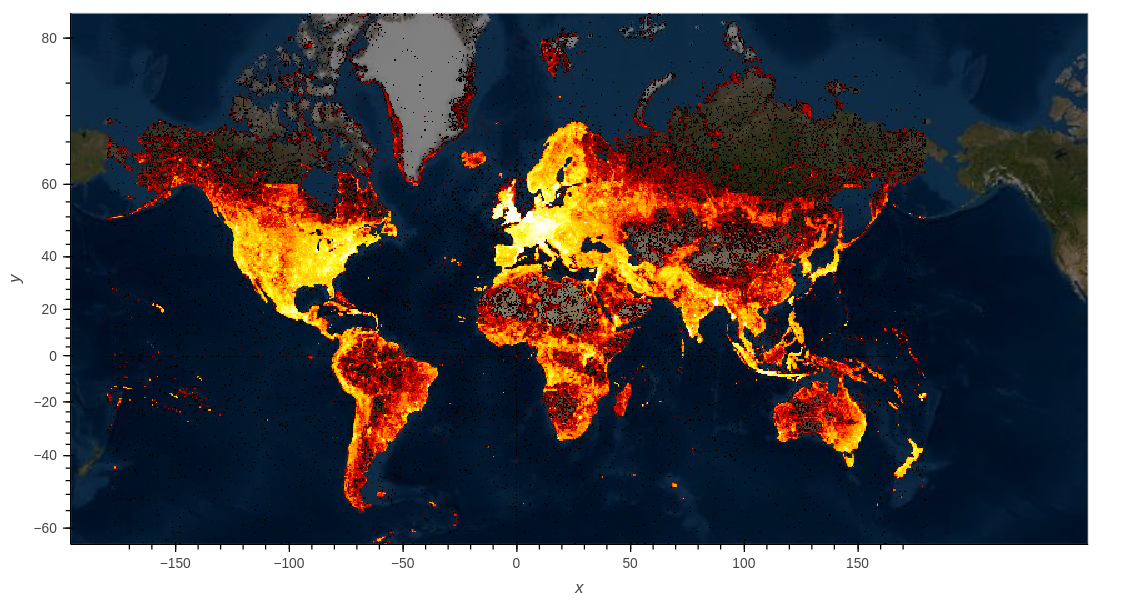}
\caption{Data points distribution.}
\label{fig:dataset}
\end{figure}

There are several reasons for choosing this subdivision scheme over a simple subdivision of latitude/longitude coordinates. Firstly, the latitude/longitude regions get elongated near the poles while S2 cells keep a close-to-quadratic shape, and secondly, S2 cells are mostly uniform in size (the ratio between the largest and smallest S2 cell is 2.08).
A naive approach to define a tiling of the earth would be to use all S2 cells at a certain fixed depth in the hierarchy, resulting in a set of roughly equally sized cells.
However, this would produce a very imbalanced class distribution since the geographical distribution of wikidata items \cite{vrandevcic} which was adopted in this paper, has strong peaks in densely populated areas see \figref{fig:dataset}.

\begin{figure} [!ht]
\centering
\includegraphics[width=12cm]{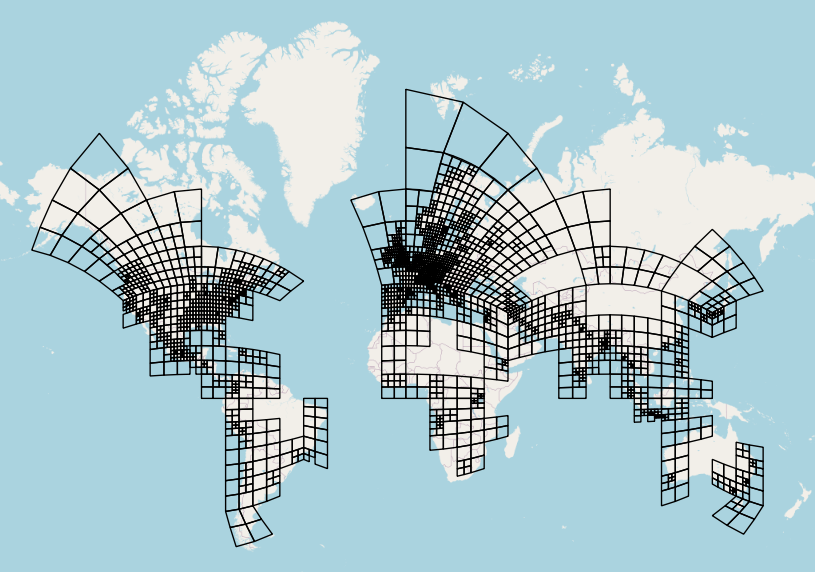}
\caption{S2 Geometry adaptive partitioning of our dataset.}
\label{fig:adaptive-partition}
\end{figure}

We therefore perform adaptive subdivision, based on the dataset item location: starting at the roots, we recursively descend each quad-tree and subdivide cells until no cell contains more than a certain fixed number (max cell samples parameter) of data points. By using this approach, sparsely populated areas are covered by larger cells and densely populated areas are covered by finer cells. This adaptive tiling has several advantages over a uniform one: (i) training classes are more balanced and (ii) it makes effective use of the parameter space because more model capacity is spent on densely populated areas. \figref{fig:adaptive-partition} demonstrates the S2 partitioning for our dataset.

\newpage
\subsection{Dataset}
This section describes the dataset used for training the model and for the evaluation experiments. 
The dataset was constructed from the wikidata archive \cite{vrandevcic2014wikidata}. 
The archive was filtered as to select all records with location label (approximately 8M records). See Tab. \ref{const_mult_dec:prob} for samples of wikidata records.

\begin{table}[!ht]
\caption{Wikipedia Geo Data samples.}
\label{const_mult_dec:prob}
\centering
\begin{tabular}{llll}
\hline
\hline
Id&Latitude&Longitude&Text\\\hline
01& 53.96& -1.08& historic county of England\\
02& 51.0& 10.0& country in Central Europe\\
03& 35.88& 14.5& sovereign state in Southern Europe\\
04& 57.30& -6.36& whisky distillery in Highland, Scotland, UK\\
05& 47.39& 0.69& city and commune in Indre-et-Loire, Centre-Val de Loire, France\\
06& -33.0& -71.0& sovereign state in South America\\
\hline
\hline
\end{tabular}
\end{table}

\subsection{Data Labeling}
\label{subsection Data Labeling}
Using the adaptive cells, we label each data sample with the cell id containing the sample location. In order to keep the hierarchical nature of the cells in the label we use the following cell id encoding: the first digit represents the cell cube face with a digit between \textit{0} to \textit{5}. The next digits represent for each level the corresponding node in the quad tree with a digit between \textit{0} to \textit{3}. See Tab. \ref{cell encoding} for a cell encoding example. 
Note that the label is a sequence of digits with a variant length between \textit{1} and \textit{1+max level}.

\begin{table}[ht!]
\caption{Cell encoding.}\label{cell encoding}
\centering
\begin{tabular}{llll}
\hline
\hline
Cell description & Cell representation&  \\\hline
Face cell 2& 2\\ 
Subcell 2 of face cell 1& 12\\
Subcell 1 of subcell 3 of face 4& 431\\
\hline
\hline
\end{tabular}
\end{table}

\section{Model}
\subsection{Train}
\begin{figure}[!ht]
\centering
\includegraphics[width=12cm]{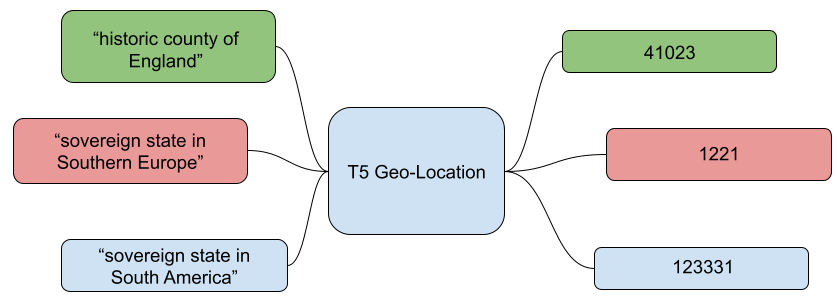}
\caption{A diagram of our text-to-location framework.}
\label{fig:training diagram}
\end{figure}

The model presented in this paper is based on the T5-base (220M) pre-trained sequence-to-sequence model \cite{raffel2020exploring}.
The model was fine-tuned on the wikidata dataset with the text records as input and the location cell encoding as the output target sequence.
We chose to utilize the standard cross-entropy loss in the fine-tuning process. Evaluation of level-based weighting in the loss function is left for future work.
We trained for 2M steps over 5 epochs with batch size of 12. We used the AdamW optimizer with learning rate 1.5e-5 and linear decay. We train the model on 80\% of the wiki-data dataset and use the other 20\% for in-domain evaluation.
A diagram of our text-to-location framework with a few input/output
examples is shown in \figref{fig:training diagram}.

\begin{figure}[!ht]
\centering
\includegraphics[width=12cm]{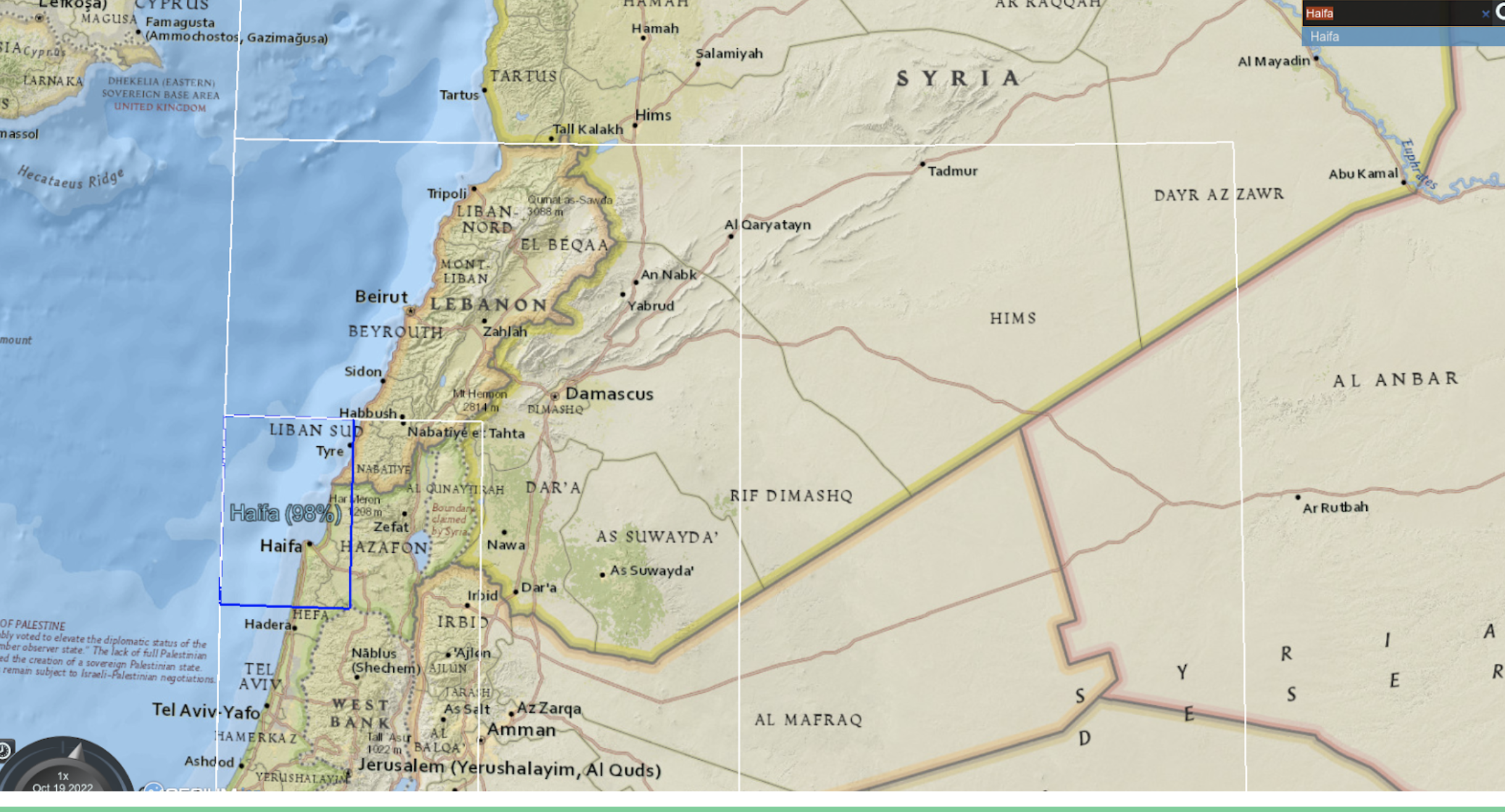}
\caption{Prediction results for the text "Haifa". The predicted S2 cell in blue and its ancestor cells in white.}
\label{fig:inference diagram}
\end{figure}
\subsection{Evaluation}
\subsubsection{Inference}
The inference of our resulting model is as follows:
given a sentence we predict the output sequence. Naturally it wold be 1 digit between 0 to 5 representing the predicted cube face, followed by up to 9 digits between 0-3 representing the predicted s2cell on each level. This sequence can easily converted to a s2 cell, which represents the probability distribution over the earth for the input text location.  
We used a beam search with size of 10. See \figref{fig:inference diagram} for inference example.

\subsubsection{Evaluation Metric}
\label{subsubEvaluation Metric}

The most straightforward classification metric is the "accuracy measure" where only predictions with a full match between the model's output and the true label counts as a successful prediction. We call this metric "Flat Accuracy" (as oppose to "Hierarchical Accuracy"). This metric fails to capture the inherent hierarchical nature of the label.
Another measure is the "Mean distance error".
Mean distance error averages the distances between the predicted location (center of the predicted S2 cell in our case) and true location of the target text. It too fails to capture the hierarchical nature of the label.

For this reason we prefer a hierarchical classification metric described in Hierarchy-Specific Variations on the Regular Classification Metrics \cite{Silla2010ASO}.
Those are variations of the well-known precision, recall and f-score metrics, specifically adapted to fit hierarchical classification:
\newline
hierarchical precision (hP):
        \begin{equation}\label{eq:1}
hP=\frac {\sum_{i} | P_{i} \cap T_{i}|} {\sum_{i} | P_{i} |},
 \end{equation}
hierarchical recall (hR):
       \begin{equation}\label{eq:2}
hR=\frac {\sum_{i} | P_{i} \cap T_{i}|} {\sum_{i} | T_{i} |},
 \end{equation}
and hierarchical f-measure (hF):
        \begin{equation}\label{eq:3}
hF=\frac {2*hP*hR} {hP+hR}
 \end{equation}
where $P_i$ is the set consisting of the most specific class predicted for each test example $i$, and all of its ancestor classes. $T_i$ is the set consisting of the true most specific class of test example $i$, and all its ancestor classes. Each summation is computed over all of the test set examples.
\newline


\section{Results}

In this section, the performance of the developed geocoding model is presented and analyzed.
To demonstrate the performance of the model, let us first present several inference examples of the model. The examples of the predicted samples are given in  Tab. \ref{example table 1}.

\begin{table}[ht!]
\caption{Inference examples - true and predicted labels.}\label{example table 1}
\centering
\begin{tabular}{llll}
\hline
\hline
Text&Predicted Label&True Label\\\hline
townland in Drummaan, County Clare, Ireland& 21002321& 21002321\\
lake in Eksjö Municipality, Sweden& 20302303&  20302303\\
ancient monument in Denmark (2976)& 20331122& 20331122\\
school in Cheshire West and Chester, UK& 210033112& 210033113\\
mountain in Iran& 1333313& 133302\\
railway stop in Harburg, Germany& 20331203& 20331022\\
\hline
\hline
\end{tabular}
\end{table}

\begin{figure}[ht!]
\centering
\includegraphics[width=12cm]{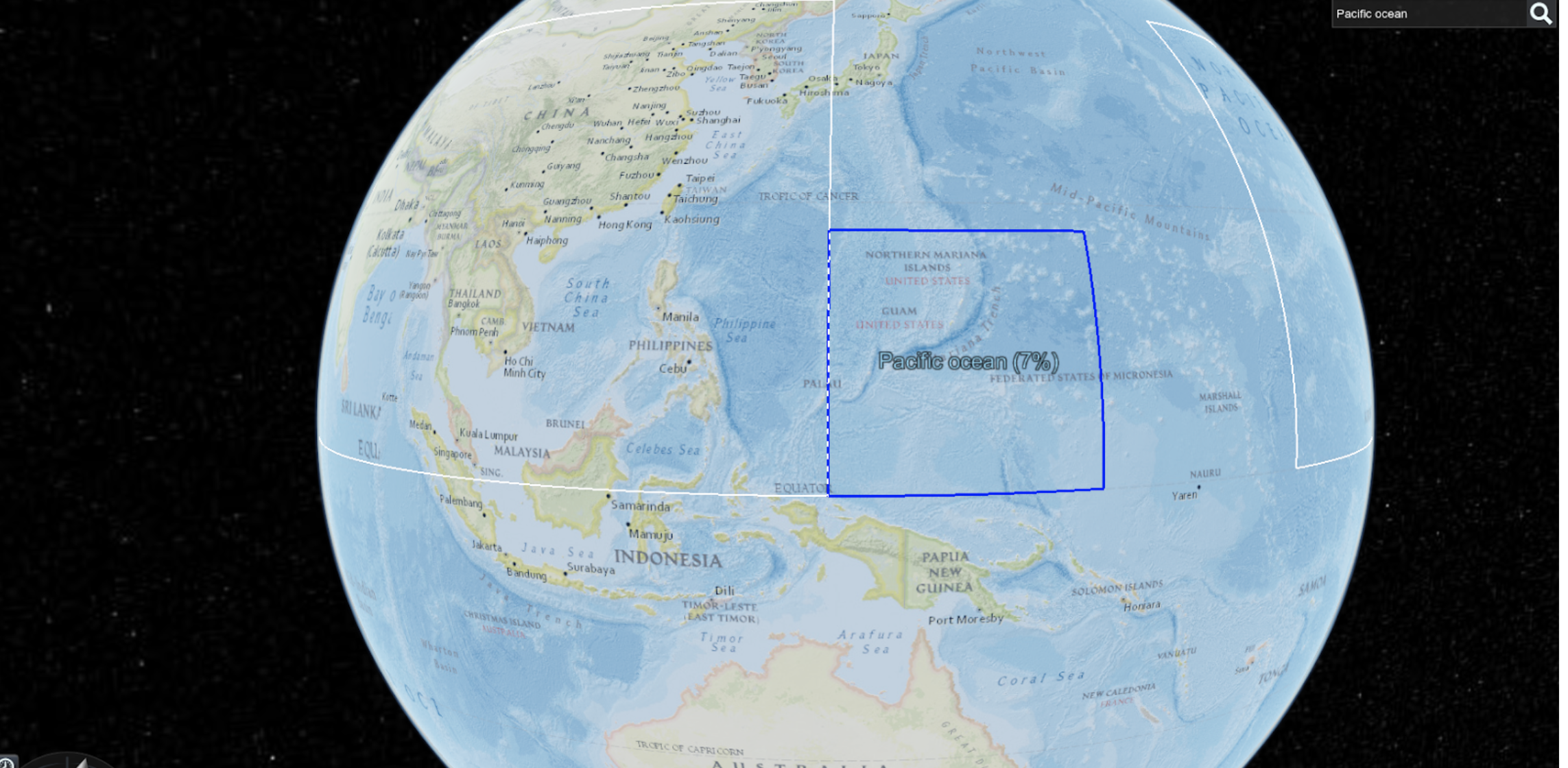}
\caption{Prediction results for the text "Pacific ocean".}
\label{fig:inference psacific}
\end{figure}

\begin{figure}[!ht]
\centering
\includegraphics[width=12cm]{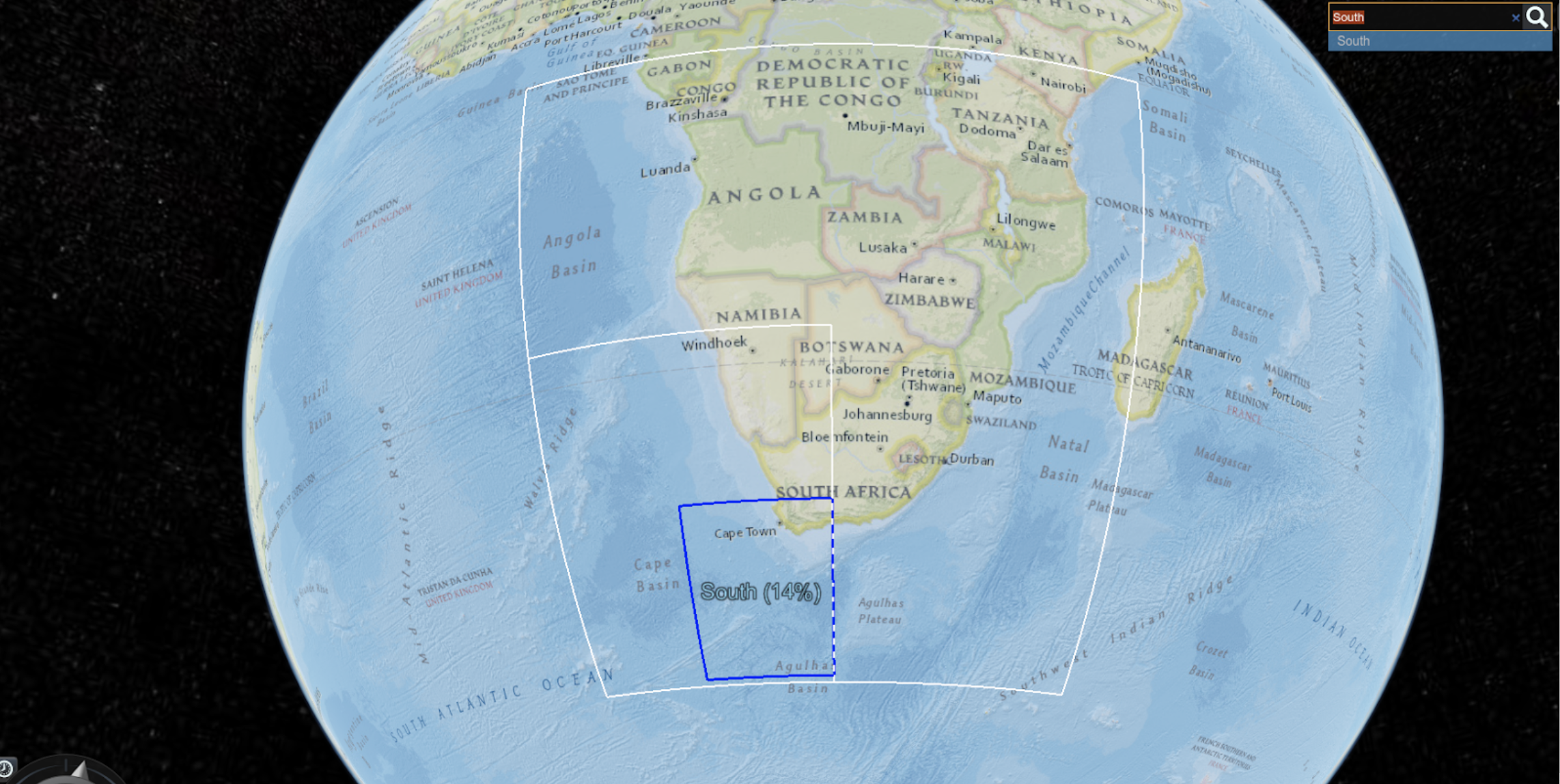}
\caption{Prediction results for the text "South". We can assume large amount of South African samples biased the prediction.}
\label{fig:inference south}
\end{figure}

\begin{figure}[!ht]
\centering
\includegraphics[width=12cm]{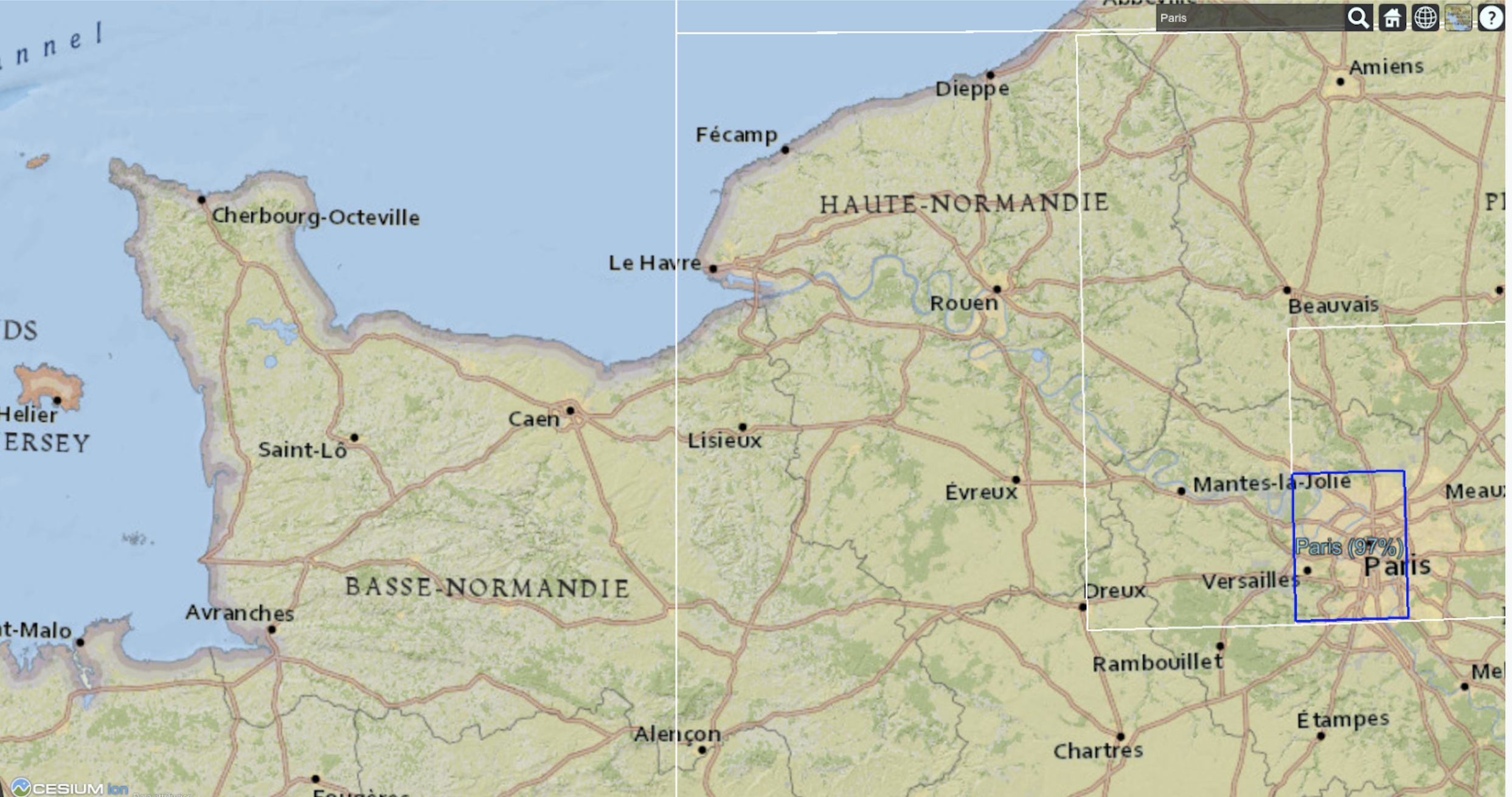}
\caption{Prediction results for the text "Paris".}
\label{fig:inference paris}
\end{figure}

It is difficult to compare two sequences due to the fact that two adjacent cells can have very different sequences in terms of string comparison.  This results from the fact that two adjacent cells can originate from different parent cells. To visualize the results, a web demo application powered by the fune-tuned model was constructed.
This demo application can facilitate an intuitive comparison between the predicted and the true label.
Inference examples using this application are given in \figref{fig:inference psacific},
\figref{fig:inference south}, and 
\figref{fig:inference paris}.

Evaluation results on the wikidata test split using the metrics described in section \ref{subsubEvaluation Metric} are given in Tab. \ref{results table}.

\begin{table}[ht!]
\caption{Evaluation results.}\label{results table}
\centering
\begin{tabular}{llll}
\hline
\hline
Evaluation metric & Results \\\hline
Flat accuracy & 0.51547\\ 
Hierarchy accuracy & 0.791\\
\hline
\hline
\end{tabular}
\end{table}

\section{Conclusions}

In this paper we formulated the problem of geocoding as a sequence-to-sequence problem and trained a transformer model for geocoding.  To leverage the capabilities of a language model, a pre-trained T5 model was used for fine-tuning. For the sequence representation of the geolocation label, an adaptive cell partitioning was used.  The free text and its corresponding geolocation were obtained from wikidata.
The evaluation of the model on both hierarchical and non-hierarchical metrics demonstrated the validity of the proposed approach for geolocation prediction.

The free text at inference time sometimes only hints at a location. It is our intuition that by combining a huge decoder such as GPT3 \cite{brown2020language} to produce more location-coherent text from an obscure location reference and then feeding it to T5 could improve the model's performance. This evaluation is left for future work.
In addition, we leave for future work the evaluation of our approach on a benchmark such as Wikipedia Toponym Retrieval \cite{gritta2018s}.

\bibliographystyle{unsrt}
\bibliography{main}

\end{document}